\documentclass{article}

\PassOptionsToPackage{numbers, compress}{natbib}

\usepackage{parskip}
\usepackage[final]{timeseries_workshop}

\usepackage[utf8]{inputenc} 
\usepackage[T1]{fontenc}    
\usepackage{hyperref}       
\usepackage{url}            
\usepackage{booktabs}       
\usepackage{amsfonts}       
\usepackage{nicefrac}       
\usepackage{microtype}      
\usepackage{xcolor}         
\usepackage{multirow}   
\usepackage{graphicx}   

\usepackage{subcaption}
\usepackage{graphicx}
\usepackage{amsmath} 
\usepackage{cleveref}

\title{\textsc{ChronoGraph}: A Real-World Graph-Based Multivariate Time Series Dataset}

\newcommand{\samethanks}[1][\value{footnote}]{\footnotemark[#1]}

\author{%
  Adrian Catalin Lutu\textsuperscript{1,2}\thanks{Equal contribution.}
  \And
  Ioana Pintilie\textsuperscript{1}\samethanks
  \And
  Elena Burceanu\textsuperscript{1}
  \And
  Andrei Manolache\textsuperscript{1,3,4}
  \AND 
  \normalfont{\textsuperscript{1}Bitdefender, Romania}\\ 
  \textsuperscript{2}University of Bucharest, Romania\\
  \textsuperscript{3}University of Stuttgart, Germany\\
  \textsuperscript{4}International Max Planck Research School for Intelligent Systems, Germany\\
  \texttt{\{alutu,ipintilie,eburceanu,amanolache\}@bitdefender.com}
}

\begin{document}

\maketitle

\begin{abstract}
We present \textsc{ChronoGraph}, a graph-structured \emph{multivariate time series forecasting} dataset built from real-world production microservices. Each node is a service that emits a multivariate stream of system-level performance metrics, capturing CPU, memory, and network usage patterns, while directed edges encode dependencies between services. The primary task is forecasting future values of these signals at the service level. In addition, \textsc{ChronoGraph} provides expert-annotated incident windows as anomaly labels, enabling evaluation of anomaly detection methods and assessment of forecast robustness during operational disruptions. Compared to existing benchmarks from industrial control systems or traffic and air-quality domains, \textsc{ChronoGraph} uniquely combines (i) multivariate time series, (ii) an explicit, machine-readable dependency graph, and (iii) anomaly labels aligned with real incidents. We report baseline results spanning forecasting models, pretrained time-series foundation models, and standard anomaly detectors. \textsc{ChronoGraph} offers a realistic benchmark for studying structure-aware forecasting and incident-aware evaluation in microservice systems. Our dataset and code are publicly available at \url{https://github.com/bit-ml/ChronoGraph}.
\end{abstract}

\section{Introduction and Related Work}

Forecasting the short- and medium-term evolution of service metrics is central to reliable operations in large-scale automated systems. Advances in connectivity, monitoring, and data collection have enabled such systems across domains ranging from manufacturing and transportation to IT infrastructure \citep{industry4}, where forecasts drive alerting, autoscaling, and capacity planning. Microservice architectures add unique challenges: hundreds of loosely coupled services form a dependency graph where disruptions (e.g., regressions, resource contention, upstream failures) can propagate across calls, making accurate forecasts dependent on both local temporal dynamics and cross-service influences. Existing graph benchmarks, however, do not reflect this setting: traffic \citep{Li2017DiffusionCR} and air-quality \citep{airqual1, geng2021trackingairpollutionchina} datasets widely used for forecasting are univariate and lack incident annotations, while industrial control datasets such as SWaT \citep{Mathur2016} and WADI \citep{Ahmed2017} include anomaly labels and are multivariate but provide only process diagrams rather than a true adjacency matrix. This leaves open the need for benchmarks that combine multivariate time series, explicit dependency structure, and labeled incidents in a single dataset.

This landscape has steered both forecasting and anomaly detection toward models that are largely topology-agnostic. Forecasters usually treat each signal independently or aggregate features without structural context. Similarly, most anomaly detection approaches operate on individual streams, with anomalies often inferred indirectly from forecast residuals or reconstruction errors \citep{Audibert2020, Zong2018, Zhang2019, Long2015,Su2019, Li2021, Zhao2020, pintilie2023diffusion, Li2019}. Foundation models for time series (e.g., Chronos, TabPFN-TS \citep{ansari2024chronos, hoo2025tabpfntimeseries}) extend forecasting capacity across domains, but they are typically applied per-series and thus cannot capture propagation effects across nodes. Existing graph-aware methods bypass actual dependencies by assuming global structures or learning dense, latent graphs end-to-end. Techniques include using attention over complete graphs \citep{mtsgraphattn}, top-$k$ similarity sampling \citep{Deng2021}, or latent graph sampling \citep{zekaietal-gta-2021}. These induced, data-driven structures are often dense and may not align with the true dependency topology.

To catalyze graph-aware forecasting in multivariate time series, we release \textsc{ChronoGraph}, a benchmark where nodes carry multivariate service metrics and edges capture call dependencies. Unlike existing datasets, it combines explicit topology with labeled incidents, enabling the evaluation of both forecasting and anomaly detection methods. This design exposes the limitations of topology-agnostic baselines and foundation models, and motivates structure-aware approaches that integrate temporal dynamics with graph information. Finally, we hope that resources like \textsc{ChronoGraph} will help connect the multivariate time series and temporal graph learning communities \citep{tempgraphs1, tempgraphs2} by providing a shared benchmark. Our contributions are as follows: 

\begin{enumerate}
\item \textbf{A real-world multivariate time series dataset.} We release \textsc{ChronoGraph}, built from six months of production microservice telemetry covering $\sim 700$ services, each represented by $5$-dimensional time series with $\sim 8{,}000$ time steps per service along with $8$-dimensional time series capturing the communication between interacting services.
    
    \item \textbf{Explicit graph topology and incident propagation.} The dataset includes a service dependency graph showing empirical anomaly propagation. This helps models capture topological correlations and supports the evaluation of topology-aware forecasting.

    \item \textbf{Baseline evaluation.} We benchmark forecasting methods, time-series foundation models, and standard anomaly detectors. We highlight key limitations of current approaches: their difficulty with long-term forecasting and their inability to leverage the system’s graph structure, and we discuss potential paths for improvement.

\end{enumerate}

\begin{figure}[t]
  \centering
  \includegraphics[width=1.\linewidth]{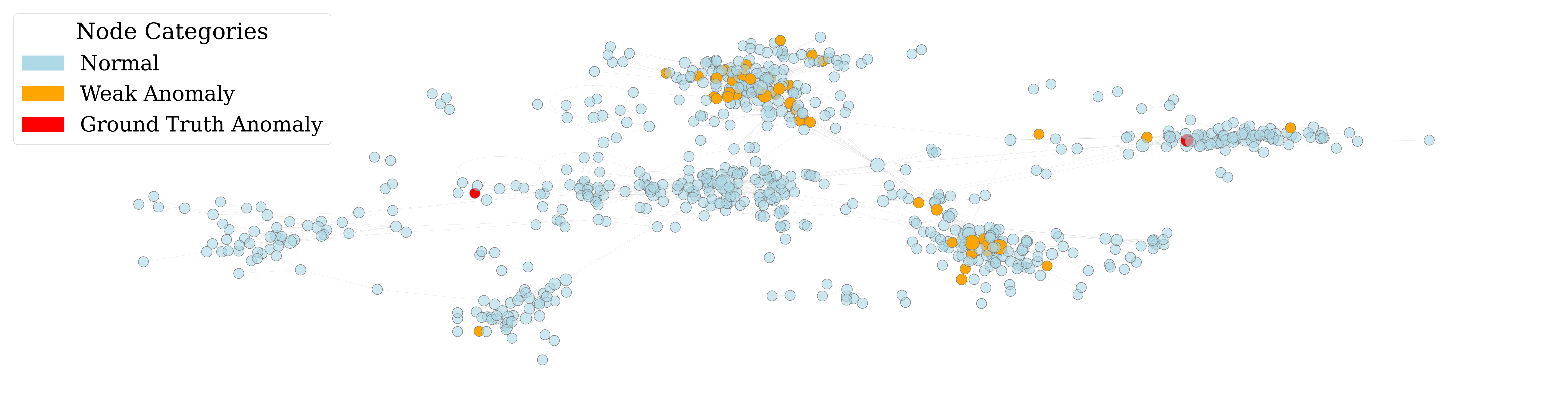 }
  \caption{System architecture at a given time. Nodes represent microservices, and edges indicate inter-service communication. Ground-truth disruptions (red) and high-confidence predictions from our ensemble (orange) are overlaid. The spatial clustering of anomalies in densely connected regions suggests that abnormal behavior tends to propagate along the underlying graph topology.}
  \label{fig:graph_propagation}
\end{figure}

\section{Data Collection and Labeling}
\noindent \paragraph{Data.}
\textsc{ChronoGraph} comprises six months of operational telemetry collected from a production microservice platform operated by a large enterprise. For each of the 708 services, we monitor five system-level metrics: CPU usage, memory usage, memory working set, and network traffic rates (incoming and outgoing). Measurements are recorded per container and aggregated to the service level (mean across containers) at a fixed 30-minute interval. After removing series with prolonged discontinuities (e.g., due to maintenance or decommissioning) and aligning timestamps, each service is represented as a multivariate time series with 8005 time steps and five variables. In addition, the observed inter-service communication defines a directed graph, where nodes are associated with multivariate time series and edges represent service dependencies along $8$-dimensions: number of requests, return codes, and latency. This explicit topology enables structure-aware forecasting and allows analysis of how disturbances propagate through the system.

\begin{table}[t]
    \centering
    \setlength{\tabcolsep}{4pt}
    \caption{Forecasting performance across models, averaged (mean$\pm$std) over all services. We report results on the full test window (3202 steps) and on the first 500 steps. All models achieve substantially lower errors on the shorter horizon, highlighting the limited long-term stability of current approaches.}
    \label{tab:model_metrics_forecasting}
    \begin{tabular}{c | c c c | c c c}
        \toprule
         & \multicolumn{3}{c|}{window=3202} & \multicolumn{3}{c}{window=500} \\
           Model   & MAE ($\downarrow$) & MSE ($\downarrow$) & MASE ($\downarrow$) & MAE ($\downarrow$) & MSE ($\downarrow$) & MASE ($\downarrow$) \\
        \midrule
         Prophet  & $\mathbf{0.125}${\tiny $\pm0.067$} & $\mathbf{0.044}${\tiny $\pm0.054$} & $7.182${\tiny $\pm11.210$} & $0.069${\tiny $\pm0.044$} & $0.013${\tiny $\pm0.022$} & $3.143${\tiny $\pm3.663$}  \\
        
         Chronos  & $0.150${\tiny $\pm0.173$} & $0.343${\tiny $\pm2.426$} & $7.902${\tiny $\pm12.707$} & $\mathbf{0.044}${\tiny $\pm0.030$} & $\mathbf{0.007}${\tiny $\pm0.015$} & $\mathbf{1.938}${\tiny $\pm1.731$} \\
        
         TabPFN-TS & $\mathbf{0.125}${\tiny $\pm0.125$} & $0.089${\tiny $\pm1.172$} & $\mathbf{6.205}${\tiny $\pm9.315$} & $0.109${\tiny $\pm0.061$} & $0.026${\tiny $\pm0.031$} & $5.082${\tiny $\pm11.013$} \\
         \bottomrule
    \end{tabular}
\end{table}

\begin{table}[t]
    \centering
    \caption{Performance of individual models and an ensemble, evaluated on human-labeled service disruptions (partial coverage). The ensemble (*) combines Prophet, Isolation Forest, and Autoencoder.}
    \label{tab:ad_metrics}
    \begin{tabular}{c c c c c c}
        \toprule
        Method & $F1_{K}$ ($\uparrow$) & $ROC_K$ ($\uparrow$) & FP rate ($\downarrow$) & FN rate ($\downarrow$) & $F1$ ($\uparrow$) \\
        \midrule
        Prophet & \textbf{20.57} & \textbf{62.97} & 2.02 & 97.98 & 2.39\\
        Isolation Forest & 17.49 & 56.39 & 46.9 & \textbf{50.48} & \textbf{7.08}\\
        OC-SVM & 14.46 & 54.31 & 22.13 & 77.08 & 5.50\\
        Autoencoder & 13.86 & 59.79 & 0.38 & 99.58 & 0.72\\
        TabPFN-TS & 12.37 & 54.08 & 0.55 & 99.79 & 0.31\\
        Chronos & 12.41 & 49.78 & 2.49 & 97.84 & 2.49\\
        Ensemble$^{*}$ & 16.92 & 60.95 & \textbf{0.20} & 99.58 & 0.73\\
        \bottomrule
    \end{tabular}
\end{table}

\noindent \paragraph{Service Disruption Labels.}
The dataset also includes a curated set of anomaly labels derived from internal incident reports. Human-written incident entries were parsed to extract affected services and timestamps, which were then mapped to fixed-length windows centered on the reported time. This procedure yields 17 labeled anomaly segments associated with specific services. 


\section{Empirical Evaluation}

\noindent \paragraph{Objectives.}
We pursue two aims: (i) \emph{Benchmark forecasting models} on \textsc{ChronoGraph}, comparing forecasting methods and time-series foundation models 
and (ii) \emph{Evaluate anomaly detection methods} on the multivariate time series using the disruption labels as ground truth. We describe the evaluation metrics, models, and experimental protocol in~\cref{apx:experiments}.

\subsection{Forecasting}
To assess the temporal stability of forecasting performance, we evaluate each model on both the full test sequence ($3202$ time steps) and a shorter horizon comprising the first $500$ steps. As shown in Table~\ref{tab:model_metrics_forecasting}, which reports average errors across all services, all models exhibit significantly lower errors in the short-term window. This consistent performance drop over longer horizons suggests that current methods lack the capacity to model long-range temporal dynamics effectively. Among them, Chronos shows the strongest decline: while it achieves the best performance on the smaller 500-point subset, it performs worst on the full sequence. In contrast, TabPFN-TS remains the most stable across both settings, and on the full sequence its performance is comparable to Prophet.

\subsection{Anomaly Detection}

The results in Table~\ref{tab:ad_metrics} show that performance remains modest across the evaluated methods. Prophet achieves the strongest individual results, though it still suffers from a very high false negative rate. Isolation Forest and OC-SVM detect more anomalies but at the cost of high false positive rates, while autoencoder-based detection and foundation models (TabPFN-TS, Chronos) perform poorly overall. The ensemble of Prophet, Isolation Forest, and Autoencoder offers somewhat more balanced performance, but remains far from reliably identifying disruption periods. Taken together, these results indicate that methods commonly used on time-series, when applied in a topology-agnostic fashion, struggle to account for the complex dynamics of microservices. Incorporating the dependency graph into forecasting or detection may offer a promising path toward more effective approaches.

It is important to stress that our ground-truth labels are sparse and limited to reported service disruptions. As a result, the dataset likely contains anomalous behaviors that were not escalated into incidents, either because they were transient or because the affected services self-recovered. While detecting these currently counts as false positives, they are operationally valuable, potentially helping operators identify issues before they become service-impacting.

\subsection{Limitations of Current Approaches}

\begin{figure}[t]
  \centering
  \includegraphics[width=1.\linewidth]{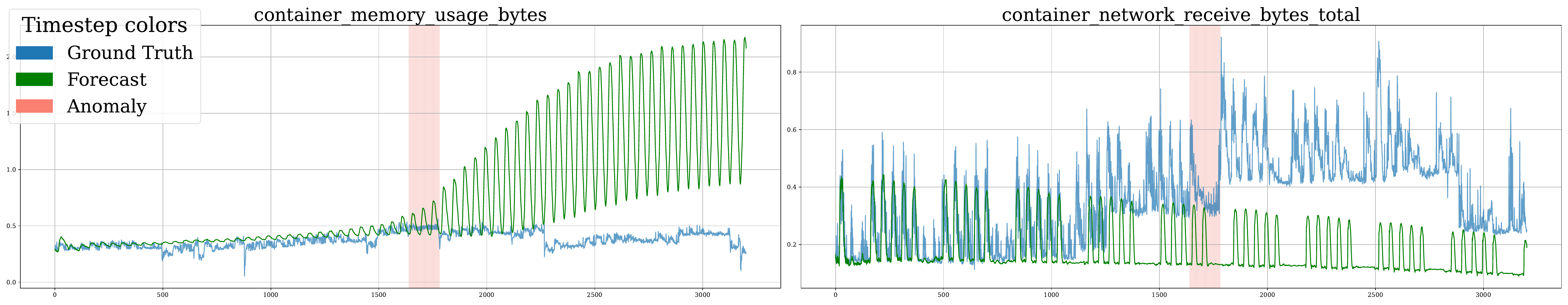 }
    \caption{Chronos forecasts on two metrics of the same microservice during a disruption. 
    \textbf{Left:} predictions drift when the anomaly occurs (red band). 
    \textbf{Right:} the model captures only a periodic baseline, missing the bursty variability of network traffic.}
  \label{fig:long_forecast1}
\end{figure}

\begin{figure}[t]
  \centering
  \includegraphics[width=1.\linewidth]{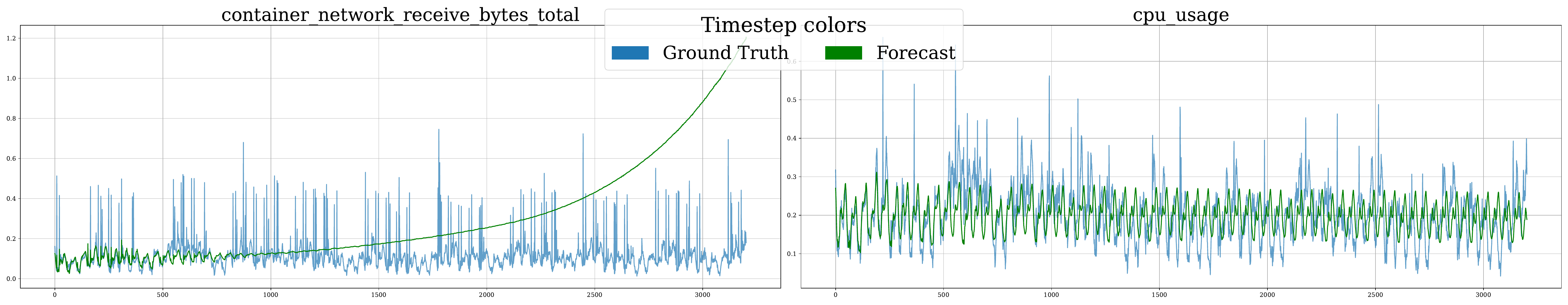 }
    \caption{Chronos forecasts on two metrics of a microservice. 
    \textbf{Left:} network traffic is accurate at first but then drifts upward even without an anomaly, illustrating the limited stability of long-horizon predictions.
    \textbf{Right:} CPU usage is forecast accurately across the full test time window.
}

  \label{fig:long_forecast2}
\end{figure}

\paragraph{Limited Forecast Horizon.}
Our results show that existing models struggle with long-term forecasting. Qualitative examples (Fig.\ref{fig:long_forecast1}, Fig.\ref{fig:long_forecast2}) illustrate that forecasts quickly diverge when the underlying series becomes volatile or when disruptions occur. Quantitatively, Table~\ref{tab:model_metrics_forecasting} shows a clear gap: while models achieve reasonable results on the first $500$ test points, performance degrades substantially when evaluated over the full horizon. This indicates that current approaches are effective at short-term prediction but lack the ability to sustain accurate forecasts as dynamics grow more complex.

\paragraph{Lack of Propagation Capabilities.}
A second limitation is that none of the models account for the service dependency graph. In Figure~\ref{fig:graph_propagation}, we plot the predictions of the system with the lowest FPR in Table~\ref{tab:ad_metrics}, which is the ensemble. Although its overall performance is limited, it combines predictions from diverse anomaly detection methods. As a result, its sparse positive predictions may reflect genuine irregularities in the series that did not escalate into labeled service failures. Predicted anomalies often cluster along connected services, suggesting that disturbances propagate through the system. These findings highlight the potential of topology-aware methods: incorporating the graph structure may improve long-term forecasting and anomaly detection by modeling both disruption spread and normal service interactions. While we do not explore such methods here, we consider them a promising direction for future work.

\section{Conclusions and Future Work}

We introduced \textsc{ChronoGraph}, a real-world dataset collected from a large-scale enterprise microservice platform. Its provenance ensures that signals and disruptions reflect real operational challenges, providing a realistic basis for model evaluation. Unlike existing benchmarks, \textsc{ChronoGraph} combines a) multivariate time series, b) a dependency graph, and c) labeled disruption windows. We benchmarked standard forecasting and anomaly detection methods across hundreds of services. Forecasting models performed well on short horizons but degraded over longer windows, revealing limited capacity to model evolving dynamics. Anomaly detectors showed modest accuracy, with predictions often clustering along connected services-suggesting propagation effects and short-lived irregularities not labeled as disruptions. These findings indicate that combining multivariate dynamics with service dependencies could improve both forecasting and anomaly detection. We present \textsc{ChronoGraph} as a foundation for advancing graph-based approaches and research in these tasks.

\section*{Acknowledgements} This project has received funding from the European Union’s Horizon  Europe  research  and  innovation  programme  under  Grant  Agreement No: 101120237 (ELIAS).

\bibliographystyle{unsrtnat}
\bibliography{bibliography}

\clearpage

\appendix

\section{Appendix}

\subsection{Further Details on the Experimental Setup}
\label{apx:experiments}

\noindent \paragraph{Metrics.} We evaluate forecasting capabilities using the mean absolute error (MAE), mean sqaured error (MSE), and the mean asbsolute scaled error (MASE). Conventional metrics for anomaly detection, such as precision, recall, and $F1$-score, can overly penalize false negatives inside a longer anomaly segment. Conversely, when combined with  point-adjustment (PA), they can severely overestimate anomaly detection performance: 
a segment is counted as correctly detected if \emph{any} point within it is flagged.
To overcome this, we adopt $F1_{K}-AUC$ and $ROC_K-AUC$ as our primary evaluation metrics. $F1_{K}-AUC$ integrates performance across different $K$ ratios of correctly predicted points \citep{kim2022towards}, providing a more balanced view of segment-level accuracy. $ROC_K-AUC$ further extends this idea by jointly evaluating across both anomaly score thresholds and $K$ values, thereby enabling fairer model comparisons regardless of hyperparameter sensitivity \citep{pintilie2023diffusion}.

\noindent \paragraph{Models.} In order to capture the strengths and weaknesses of fundamentally different modeling paradigms, we choose a broad set of approaches spanning statistical models, representation learning, foundation models, and classical machine learning methods. We include two time series foundation models: Chronos-Bolt Base \citep{ansari2023chronos}, designed for zero-shot and few-shot time series tasks, and TabPFN-TS \citep{hoo2025tabpfntimeseries}, a transformer-based prior-data-fitted network adapted for time series. Within the class of statistical forecasting methods, we employ Prophet \citep{taylor2018forecasting}, a decomposable model for trend, seasonality, and holiday effects, since ARIMA~\citep{box2015time} in our setting mostly defaulted to forecasting the mean under noisy conditions. For anomaly detection, in addition to using residuals from the forecasting models, we also consider Autoencoders \citep{sakurada2014anomaly}, which detect anomalies through reconstruction error, along with two classical AD methods, Isolation Forest \citep{liu2008isolation} and the One-Class SVM (OC-SVM) \citep{scholkopf2001estimating}, as they remain strong baselines for high-dimensional outlier detection.

\noindent \paragraph{Training details.} We apply a 60/40 train–test split to ensure sufficient history for model learning while retaining ample data for evaluation. For Prophet, we train a separate model for each dimension to capture individual temporal dynamics. For Chronos, we adopt a rolling prediction strategy: we forecast the first 64 time steps, append them to the context, and then use this extended context to predict the next 64 time steps, repeating the process until the forecast horizon is reached. In contrast, TabPFN-TS is trained jointly on the entire MTS, leveraging cross-series dependencies to model shared temporal patterns. The resulting forecasts from all models are then used for anomaly detection by computing z-scores of the residuals, which are subsequently evaluated using the anomaly detection protocol described in \citep{pintilie2023diffusion}

\subsection{Additional Figures}





\begin{figure}[t]
  \centering
  \includegraphics[width=1.\linewidth]{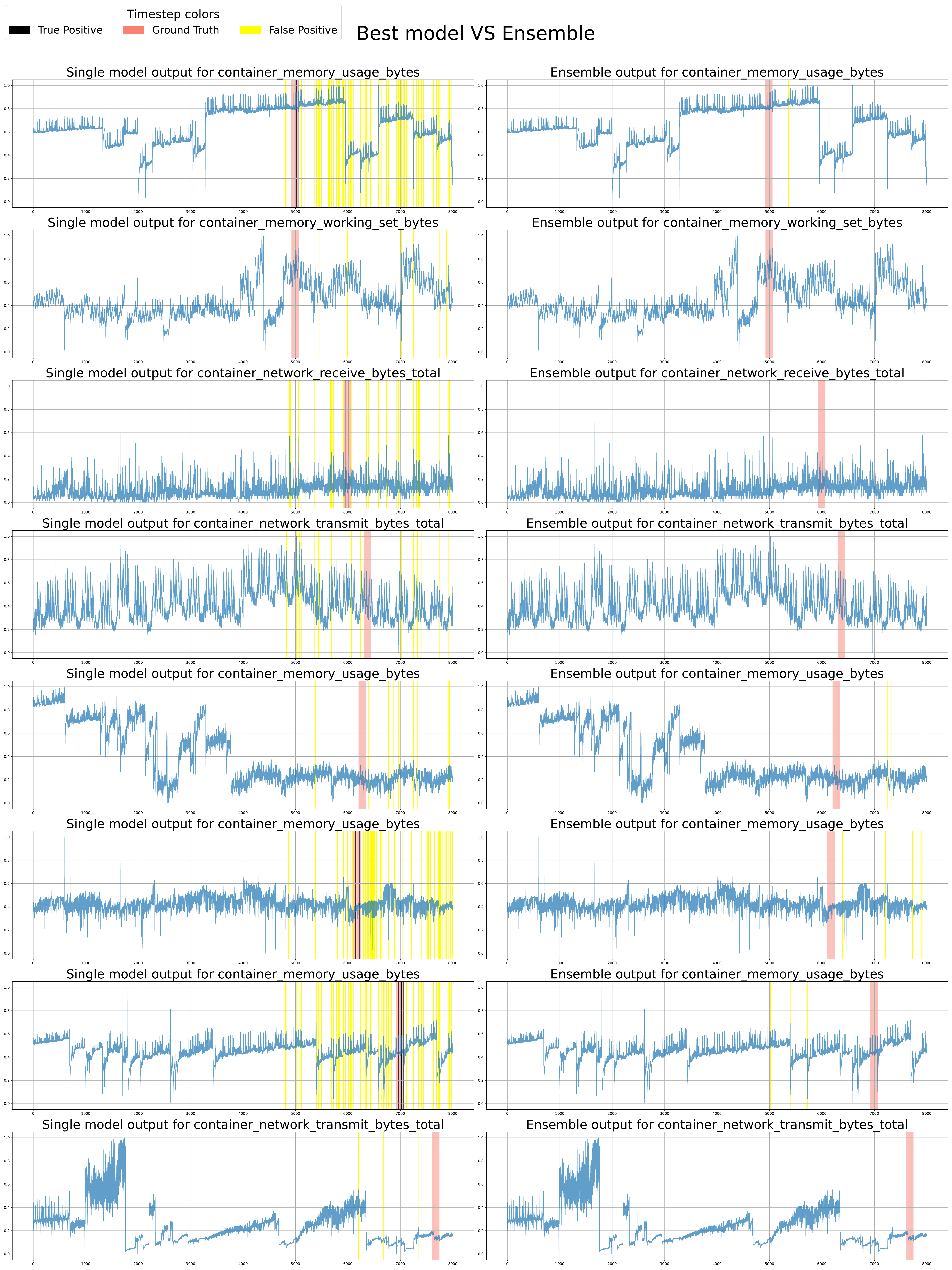 }
  \caption{Comparison of Prophet model and ensemble anomaly detection outputs across multiple container metrics}
  \label{fig:ensemble_ap}
\end{figure}
\begin{figure}[t]
  \centering
  \begin{subfigure}[t]{0.48\linewidth}
    \centering
    \includegraphics[width=\linewidth]{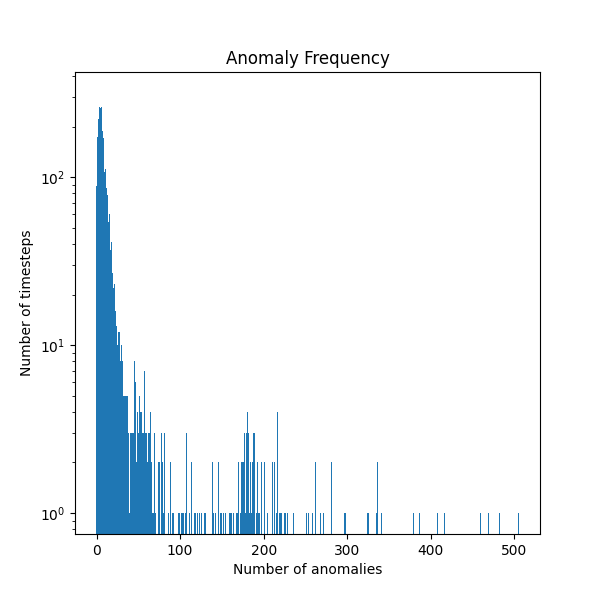}
    \caption{Anomaly Frequency Prophet}
    \label{fig:anomaly_freq_prophet}
  \end{subfigure}
  \hfill
  \begin{subfigure}[t]{0.48\linewidth}
    \centering
    \includegraphics[width=\linewidth]{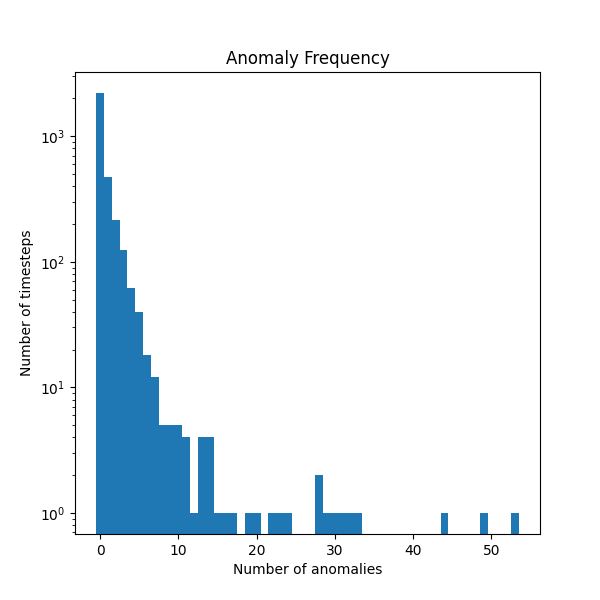}
    \caption{Anomaly Frequency Ensemble}
    \label{fig:anomaly_freq_ensemble}
  \end{subfigure}
  \caption{Comparison of anomaly frequency distributions.}
  \label{fig:anomaly_freq_sidebyside}
\end{figure}
\begin{figure}[t]
  \centering
  \includegraphics[width=1.\linewidth]{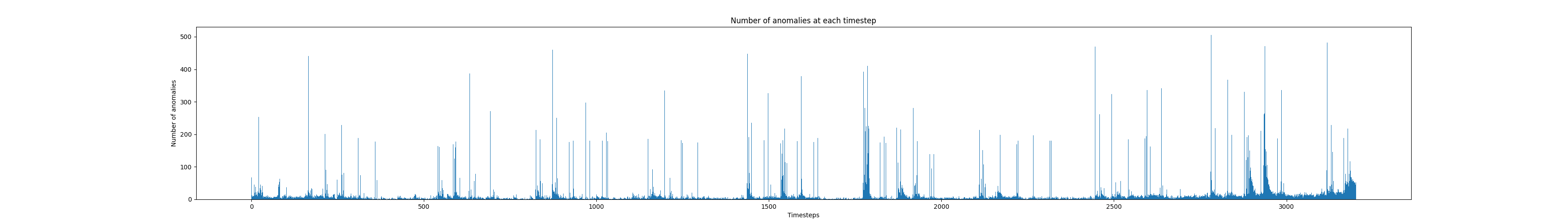 }
  \caption{Anomalies at each time step found by Prophet}
  \label{fig:anomalies_ts_prophet}
\end{figure}
\begin{figure}[t]
  \centering
  \includegraphics[width=1.\linewidth]{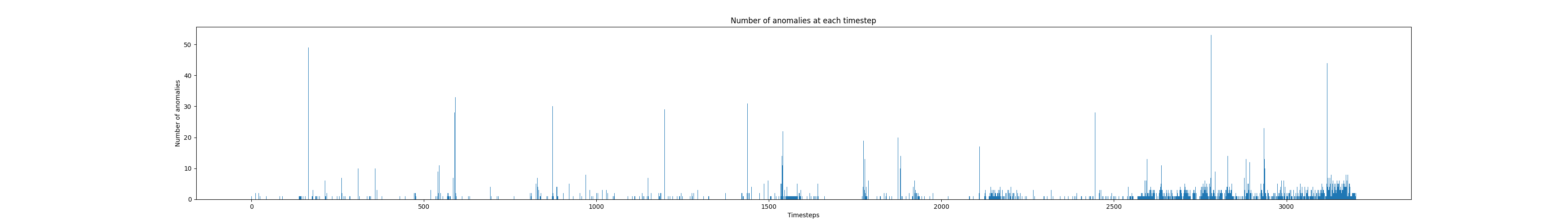 }
  \caption{Anomalies at each time step found by the ensemble}
  \label{fig:anomalies_ts_ensemble}
\end{figure}

Figure~\ref{fig:ensemble_ap} highlights anomaly detection qualitative examples for both Prophet and the ensemble model. While Prophet successfully detects multiple points inside the ground truth anomaly segment, it has a higher FPR than the ensemble. Figure~\ref{fig:anomaly_freq_sidebyside} further compares the distributions of anomaly counts per time step detected by the ensemble model versus the Prophet model. While most time steps contain few or no anomalies, a small subset exhibit higher anomaly counts, suggesting that anomalies may propagate along the graph structure. The y-axis is log-scaled, highlighting the heavy-tailed nature of anomaly occurrence, where rare but extreme anomaly events are present. Finally, Figures~\ref{fig:anomalies_ts_prophet} and~\ref{fig:anomalies_ts_ensemble} present the distributions of anomalies per time step identified by the Prophet model and the ensemble approach, respectively.

\end{document}